\documentclass[letterpaper, 10 pt, conference]{ieeeconf}  

\IEEEoverridecommandlockouts                              

\usepackage{amsmath} 
\usepackage{amssymb}  

\usepackage{tabularx}
\usepackage{svg}
\usepackage{booktabs}
\usepackage{subcaption}
\usepackage{multirow}
\usepackage[symbol]{parnotes}
\usepackage{siunitx}
\usepackage{nicematrix}
\usepackage{placeins}

\title{\LARGE \bf
HYPE: Hybrid Planning with Ego Proposal-Conditioned Predictions\\
}

\author{Hang Yu$^{1,2\dag}$, Julian Jordan$^{1}$, Julian Schmidt$^{1}$, Silvan Lindner$^{1}$, Alessandro Canevaro$^{1,3}$, and Wilhelm Stork$^{2}$
\thanks{*This work is a result of the joint research project "STADT:up" (19A22006O)", supported by the German Federal Ministry for Economic Affairs and Climate Action (BMWK), based decisions of the German Bundestag. The author is solely responsible for the content of this publication.
}
\thanks{$^{1}$Mercedes-Benz AG, Research \& Development, Stuttgart, Germany}%
\thanks{$^{2}$Karlsruhe Institute of Technology, ITIV, Karlsruhe, Germany}%
\thanks{$^{3}$University of T{\"u}bingen, T{\"u}bingen, Germany}%
\thanks{$^{\dag}${\tt\small hang.yu@mercedes-benz.com}}%
}

\begin{document}

\maketitle
\thispagestyle{empty}
\pagestyle{empty}

\begin{abstract}
Safe and interpretable motion planning in complex urban environments needs to reason about bidirectional multi-agent interactions. This reasoning requires to estimate the costs of potential ego driving maneuvers.
Many existing planners generate initial trajectories with sampling-based methods and refine them by optimizing on learned predictions of future environment states, which requires a cost function that encodes the desired vehicle behavior.
Designing such a cost function can be very challenging, especially if a wide range of complex urban scenarios has to be considered. 
We propose HYPE: HYbrid Planning with Ego proposal-conditioned predictions, a planner that integrates multimodal trajectory proposals from a learned proposal model as heuristic priors into a Monte Carlo Tree Search (MCTS) refinement. To model bidirectional interactions, we introduce an ego-conditioned occupancy prediction model, enabling consistent, scene-aware reasoning. Our design significantly simplifies cost function design in refinement by considering proposal-driven guidance, requiring only minimalistic grid-based cost terms.
Evaluations on large-scale real-world benchmarks nuPlan and DeepUrban show that HYPE effectively achieves state-of-the-art performance, especially in safety and adaptability.

\end{abstract}

\section{Introduction}
\label{sec:intro}
Motion planning for Autonomous Vehicles (AVs) demands robust reasoning about dynamic interactions within complex, uncertain driving environments. In real-world traffic, surrounding agents exhibit diverse and interactive behaviors influenced by individual objectives, environmental context, and social dynamics. To generate safe and efficient trajectories, a planner needs to anticipate how the ego vehicle's actions influence and are influenced by other agents. Methods that do not effectively capture these interactions often result in overly conservative or unsafe trajectories \cite{Chen2023TreestructuredPP}.

Traditional motion planners typically fall into three categories. Purely rule-based systems \cite{fan2018baiduapolloemmotion, Treiber_2000, Dauner2023CORL} are interpretable and provide safety guarantees, but require extensive manual adjustment of the cost function. Fully learned approaches such as imitation learning methods~\cite{scheel2021urban, bansal2018chauffeurnet, vitelli2021safetynet, hallgarten2023prediction} minimize manual intervention but often struggle with interpretability, safety guarantees, and generalization to unforeseen scenarios \cite{karkus2022diffstack}. Hybrid planners bridge rule-based and learning-based methods by using data-driven modules to generate or score trajectory proposals within a classical planning framework, thus retaining safety guarantees and improving adaptability.

\begin{figure}[t]
  \centering
  \includegraphics[width=\linewidth]{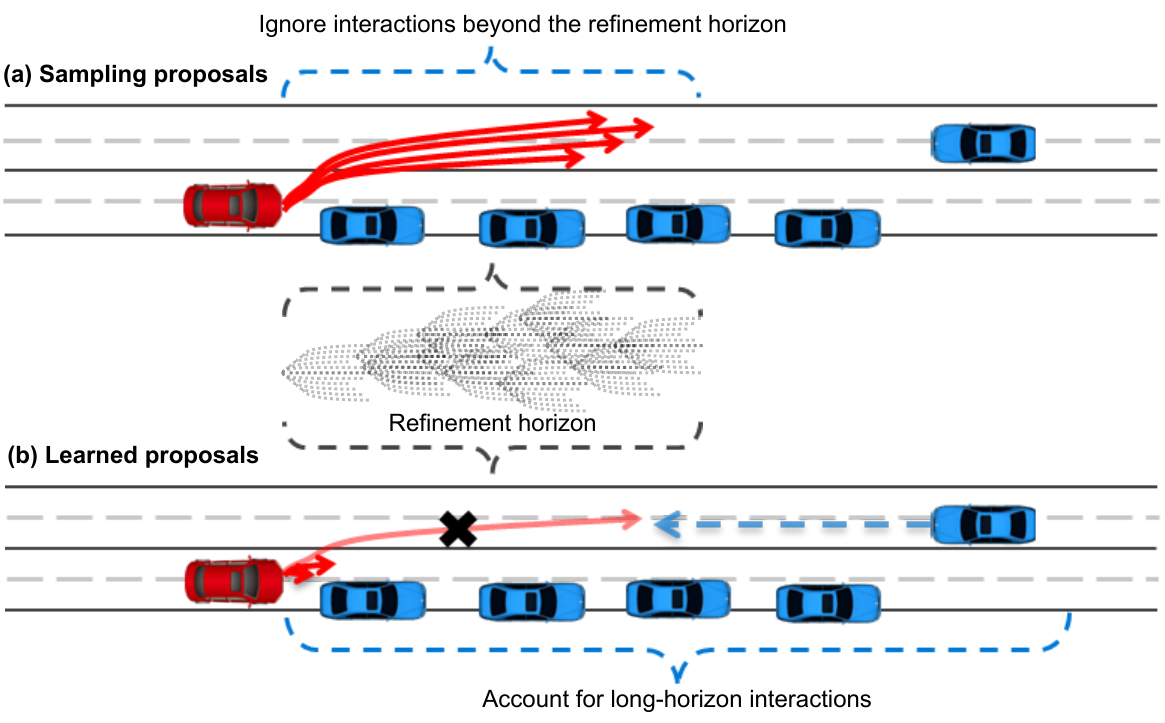}
  \caption{Different proposal strategies in the hybrid planners. 
    (a) Existing sampling proposals are subject to a given optimization horizon.
    (b) Our proposed approach employs a learnable multimodal ego proposal network to generate flexible trajectories that capture long-term scene interactions.}
  \label{fig:intro}
\end{figure}

Most recent hybrid planners integrate learned predictions of future scene dynamics with rule-based optimization to refine initial trajectory proposals, as in works such as~\cite{Chen2023TreestructuredPP, huang2024dtpp}, which acknowledge interactions between the ego and surrounding agents.
However, as illustrated in Fig.~\ref{fig:intro} their primary proposals typically rely on sampling heuristics such as lane centerlines and fixed velocity targets, and the sampling horizon is limited by computational complexity of the downstream refinement process. These proposals often lack sufficient diversity and fail to consider medium- and long-term scene interactions, leading to trajectories trapped in local minima during refinement. Furthermore, extensive fine-tuning of the cost functions is required for various scenarios.

To address these limitations, we propose HYPE, a hybrid planning approach that integrates multimodal ego proposals from a state-of-the-art (SOTA) proposal model with a structured heuristic search. The method leverages proposal-guided planning, where multimodal ego trajectories inform the Monte Carlo Tree Search (MCTS), providing strong priors that guide the exploration process. These proposals are further used to condition occupancy prediction, allowing the planner to anticipate scene evolution in a manner that is both interaction-aware and consistent with the ego's potential future maneuver. Finally, we adopt a lightweight grid-based cost function that focuses exclusively on collision avoidance and proposal adherence, thereby reducing the reliance on manually tuned cost terms while ensuring driving safety. Our approach is extensively validated on two large-scale real-world datasets, nuPlan \cite{nuplan} and DeepUrban \cite{selzer2024deepurban}, demonstrating superior safety performance and adaptability compared to SOTA baselines. The results confirm our method's capability to consistently produce safe, interpretable, and interaction-aware plans suitable for complex urban driving scenarios.

In summary, our key contributions are as follows:
\begin{itemize}
    \item We introduce a hybrid planning approach that integrates multimodal learned ego proposals and structured heuristic search, effectively balancing learned guidance with rule-based safety constraints.
    \item Our occupancy prediction explicitly conditioned on ego proposals provides scene-consistent and interaction-aware forecasts, enhancing planning safety and responsiveness.
    \item By implicitly encoding complex behavioral reasoning within learned proposals, our approach simplifies cost functions to essential safety checks, substantially reducing manual tuning and improving generalization across diverse scenarios.
\end{itemize}

\section{Related Work}

\subsection{Optimization-Based Planning:}
Optimization-based planning has long been a cornerstone in autonomous driving, typically relying on handcrafted cost functions and rule-based strategies to generate safe trajectories \cite{fan2018baiduapolloemmotion, Treiber_2000, Dauner2023CORL}. Classic search-based methods such as A* \cite{Search-Based_Optimal} and dynamic programming \cite{fan2018baiduapolloemmotion} offer interpretability and control, but require extensive manual tuning of cost terms and weights to handle the diverse conditions encountered in real-world driving. Furthermore, when modeling interactions among multiple agents, these approaches can become computationally inefficient and are prone to becoming trapped in local minima.

MCTS \cite{chaslot_monte-carlo_2021} has been utilized as an effective search-based technique in planning \cite{Kurzer_2018, Game-Theoretic_Strategy} due to its ability to balance exploration and exploitation in large, non-convex spaces without relying on overly restrictive domain assumptions. However, standard MCTS implementations often incur high computational costs when evaluating numerous candidate trajectories.

To generate scene-aware ego trajectory proposals that effectively guide the tree expansion process, we extend the established optimization-based frameworks by integrating a proposal heuristic into the MCTS architecture \cite{heuristic_MCTS}. 

\subsection{Data-Driven Planning:}
Data-driven planning methods have often adopted imitation learning, such as behavior cloning, to model human driving behaviors directly from raw sensor data \cite{Chitta2023TransFuser, shao2022interfuser, jia2023thinktwice, CodevillaBehaviorCloning}. Although these approaches can produce human-like responses, they yield outputs that lack explicit safety guarantees and are challenging to fine-tune across varying driving conditions. In contrast to operating directly on raw sensor data, other related work, such as \cite{NMP, hu2022stp3}, uses interpretable intermediate representations.
They have introduced a spatial-temporal cost volume technique, which provides enhanced interpretability and naturally captures uncertainty. However, these methods rely exclusively on neural network components, which can hinder the integration of principled planning and control algorithms common in practice \cite{karkus2022diffstack}.

Our work combines the strengths of a spatial-temporal cost volume technique with a tree-structured planning framework. This integrated approach offers an interpretable and safety-aware solution that effectively balances data-driven proposals with structured, rule-based safety constraints.

\subsection{Joint Prediction and Planning:}
Joint prediction and planning are crucial in highly interactive driving scenarios \cite{hagedorn2024integrationpredictionplanningdeep}.
Several studies have proposed holistic neural network approaches to jointly generate trajectories for both the ego and other agents \cite{scheel2021urban, karkus2022diffstack, Huang_Differentiable_PRED_PLAN}. However, these methods overlook the bidirectional influences between the ego vehicle and the surrounding agents.

Works such as \cite{Chen2023TreestructuredPP, huang2024dtpp, Huang_2023_ICCV, song2020pip, Tractable_Rhinehart} incorporate ego-conditioning by tailoring predictions to the ego’s future proposal. However, many of these approaches rely on basic sampling strategies that generate trajectory proposals based on lane centerlines and discrete velocity or acceleration candidates, frequently in combination with object-based prediction.
This limits trajectory proposals to a scope defined by the pre-planned route and discards valuable scene dynamic information. 
Moreover, limiting predictions to a fixed number of neighboring agents further exacerbates the challenge of navigating complex and crowded urban scenarios.


Our approach combines an ego proposal heuristic planner with ego-conditioned spatio-temporal occupancy grids. 
A learnable proposal network incorporates long-term scene dynamics as guidance, while grid-based cost evaluation provides ego-aware cost volumes for safe and efficient planning.

\begin{figure*}[htbp]
    \centering
    \includegraphics[width=\textwidth]{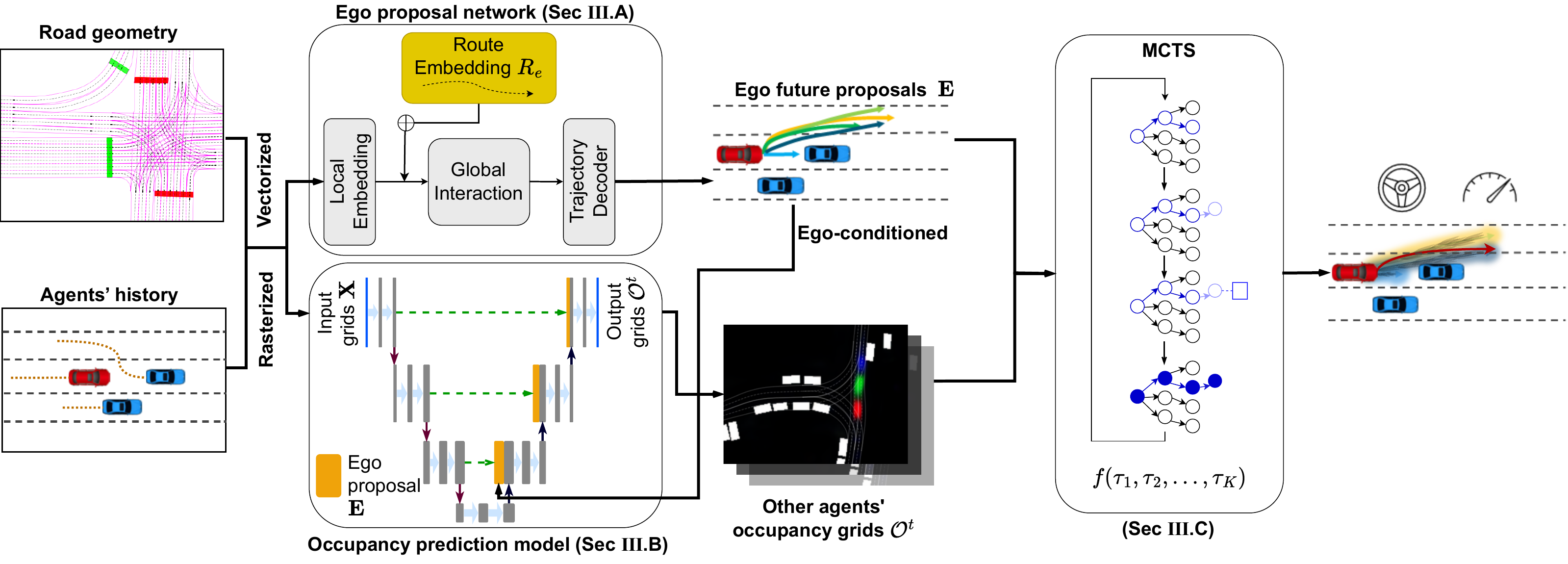}
    \caption{Model architecture of the hybrid heuristic planner HYPE. The model integrates an ego proposal network that explicitly encodes route embeddings to generate multimodal ego future proposals, which condition the occupancy prediction and guide the subsequent MCTS refinement.}
    \vspace{-0.5em}
    \label{fig:model_architecture}
\end{figure*}

\section{Methodology}
In this section, we present HYPE, the model architecture depicted in Fig.~\ref{fig:model_architecture}.
Specifically, we enhance an ego proposal network by incorporating explicit route embeddings, enabling it to generate trajectory proposals aligned with the intended navigation paths. Additionally, we introduce an ego-conditioned occupancy prediction model, allowing occupancy predictions to dynamically respond to the ego vehicle's planned maneuver. Finally, our MCTS planner systematically explores and simulates these proposals using a structured heuristic-guided exploration strategy, coupled with a grid-based convolutional cost function. 

\subsection{Ego Proposal Network}
\label{sec:ego_proposal}
To generate trajectory proposals for the ego vehicle, we build upon the multimodal trajectory prediction framework HiVT \cite{zhou_hivt_2022}, augmented by explicit route guidance. Our approach extracts discrete centerlines as the navigation route on the lane graph for the ego vehicle.

We encode these route waypoints \((x_i,y_i,\theta_i)\) for the length \(i=1,\ldots,L\) into the embedding \(R_{e}\in\mathbb{R}^{L\times E}\) using an MLP. 
To effectively integrate these route embeddings with the ego vehicle's representation from the local encoder, we employ multi-head attention (MHA). The resulting enhanced ego representation is then obtained by fusing these attended features with the original ego state through a two-layer MLP.
\begin{equation}
    F'_{\mathrm{ego}} = \mathrm{MLP}_{f}\bigl([\;F_{\mathrm{ego}},\;\mathrm{MHA}(F_{\mathrm{ego}},R_{e},R_{e})\;]\bigr)\in\mathbb{R}^{E}.
\end{equation}
By conditioning ego proposals on navigation routes, the proposal network generates context-aware ego future proposals. 

\subsection{Ego-Conditioned Occupancy Prediction}
\label{sec:occ_pred}
We adopt a U-Net \cite{ronneberger_u-net_2015} based model to predict spatio-temporal occupancy grids that encode the occupancy probabilities of surrounding agents at future time steps. To explicitly condition these predictions on the ego vehicle's planned maneuver, we integrate ego future trajectory proposals $\mathbf{E}$ into the decoder stages, distinguishing our approach from a standard U-Net.

To construct the input, all representations follow the ego-centric rasterization convention. We construct the input \(\mathbf{X}\) from three components.

\noindent\textbf{Pose History $\mathbf{P}$}: A stack of historical agent poses (including the ego), captured over the past $T_{\text{pose}}$ timesteps. These poses are rasterized into binary occupancy grids $\mathbf{P} \in \mathbb{R}^{D_{\text{pose}} \times T_{\text{pose}} \times H \times W}$, where $D_{\text{pose}}$ separates channels for the ego and other agents, and $T_{\text{pose}}$ denotes historical frames.

\noindent\textbf{Static World $\mathbf{S}$}: A multi-channel grid that encodes static road geometry, including lane directions, drivable areas, and lane markings, and is represented as $\mathbf{S} \in \mathbb{R}^{D_{\text{static}} \times H \times W}$.

\noindent\textbf{Ego Future Proposal $\mathbf{E}$}: A binary grid encoding the ego vehicle’s proposed future states over $T_{\text{future}}$ timesteps: $\mathbf{E} \in \mathbb{R}^{T_{\text{future}} \times H \times W}$.


During decoding, to ensure that occupancy predictions align with the intended trajectory of the ego, we condition feature upsampling on the ego future proposal, as depicted in Fig.~\ref{fig:model_architecture}.

\subsection{Monte Carlo Tree Search Refinement}
\label{sec:mcts}
MCTS is a widely used decision-making algorithm that iteratively refines action selection through structured exploration. Each iteration consists of four key steps: selection, expansion, simulation, and backpropagation \cite{chaslot_monte-carlo_2021}.
Unlike the vanilla MCTS, we integrate learned ego proposals into expansion exploration and apply a light-weight and general grid-based cost function in simulation to select the child node with the highest accumulated reward at each iteration.  


\subsubsection{Proposal Heuristic Exploration}
\begin{figure}[t]
  \centering
  \includegraphics[width=\linewidth]{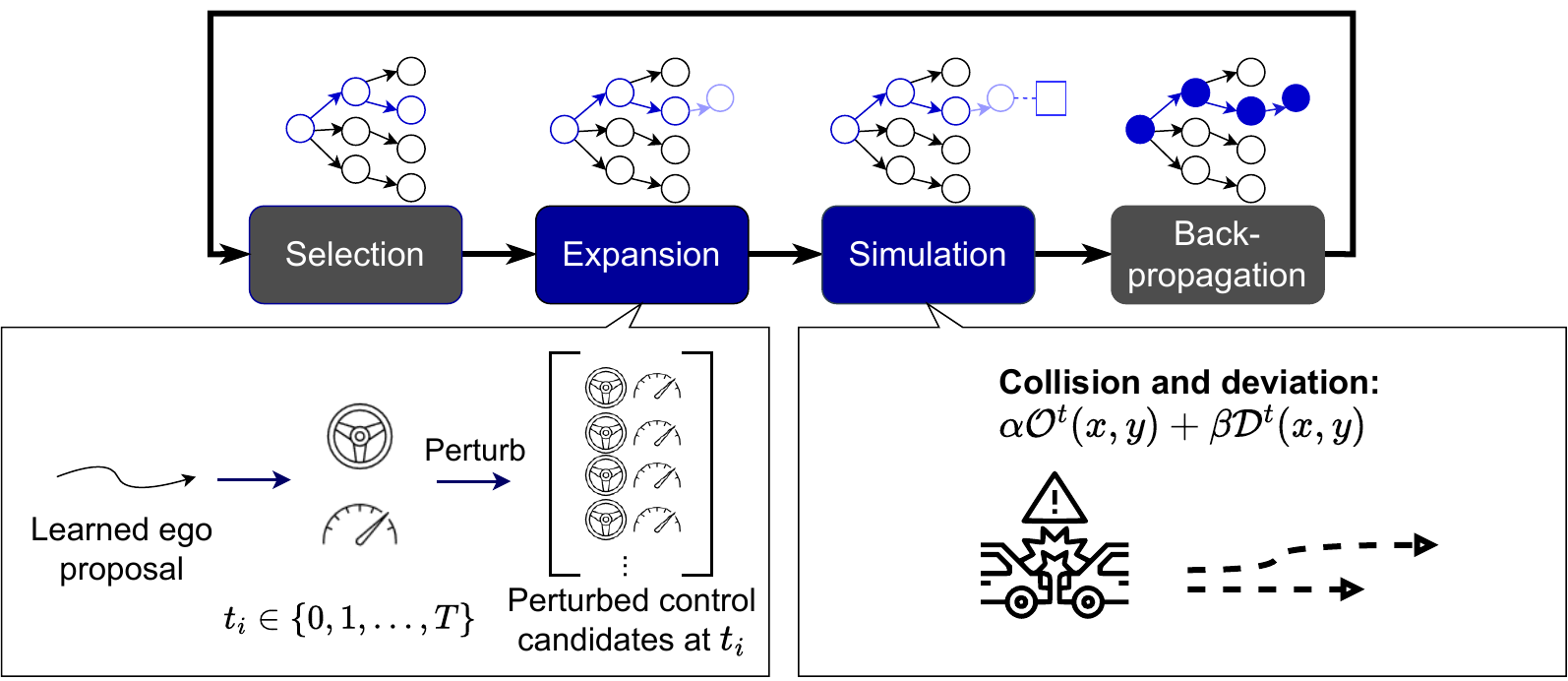}
  \vspace{-1.5em} 
  \caption{Proposal-guided MCTS planning. During expansion, nodes are chosen from the set of perturbed proposal candidates. In simulation, each rollout is evaluated using a lightweight cost function combining collision risk and proposal deviation.}
  \vspace{-1.0em}
  \label{fig:guided_mcts}
\end{figure}

As illustrated in Fig.~\ref{fig:guided_mcts}, our method employs MCTS to build and explore a search tree by expanding feasible ego proposals.
During selection, it recursively traverses the tree, choosing the most promising child based on the Upper Confidence Bound (UCB) policy:
\begin{equation}
\underset{n_i \in \text{children}(n_t)}{\arg\max} \left[
\frac{Q(n_i)}{V(n_i)} + c \sqrt{\frac{ \ln V(n_t)}{V(n_i)}}
\right],
\label{eq:ucb_selection}
\end{equation}
where \( Q(n_i) \) is the total reward of child node \( n_i \), \( V(n_i) \) and \( V(n_t) \) are the visit counts of the child and parent nodes, and \( c \) is an exploration constant, typically set to \( \sqrt{2} \).

Unlike uninformed search methods \cite{Kurzer_2018, Game-Theoretic_Strategy}, we adopt learned ego trajectory proposals as heuristic guidance to improve search efficiency. 
During expansion, child nodes are generated by discretely perturbing acceleration and steering around each of the $K$ ego proposals, enabling local exploration in action space.  
To control computational complexity, we apply progressive widening to limit the number of children per node by its visit count:
\begin{equation}
N_{\text{expand}} = \lfloor k \cdot v_n^{\gamma} \rfloor,
\label{eq:progressive_widening}
\end{equation}
where $v_n$ is the node's visit count, and $k$, $\gamma$ are hyperparameters controlling the expansion rate. 
Each expanded node state is propagated using a kinematic bicycle model \(s_{n_i}=f_{\text{bicycle}}(s_{n_t},a,\delta,\Delta t)\), ensuring realistic dynamics.

\subsubsection{Grid-Based Convolutional Cost Evaluation}
The simulation phase involves running a rollout from the newly expanded node to estimate its future rewards.  
Trajectory evaluation uses a minimalistic grid-based convolutional cost function that combines predicted occupancy distribution and deviation from the proposal:
\begin{equation}
C(\tau) = \max_{t=1}^{T} \sum_{x,y} W_{\text{ego}}(x,y) * \left( \alpha \mathcal{O}^t(x,y) + \beta \mathcal{D}^t(x,y) \right),
\label{eq:cost_function}
\end{equation}
where $\alpha$ and $\beta$ balance the penalties for the predicted collision likelihood and proposal deviation. For each time step $t$, the cost is calculated by convolving a fixed ego-shaped kernel $W_{\text{ego}}$ over a weighted combination of the occupancy grid $\mathcal{O}^t$ and the deviation map $\mathcal{D}^t$.
This process forms a spatio-temporal cost volume along the trajectory, and the final cost is defined as the maximum value over time, prioritizing the most critical moment for safe decision making. By encoding complex behaviors within learned proposals and evaluating only essential safety constraints, this formulation simplifies and generalizes the cost function across diverse planning scenarios.

\subsubsection{Structured Rollout Policy}
During simulation, the controls derived from the proposals are slightly perturbed by Gaussian noise \(\mathcal{N}(0,\sigma^2)\), clipped within physical actuation limits, and then propagated forward using the bicycle model.
A smoothness constraint (maximum allowable jerk) ensures realistic trajectory behavior.

Finally, in the backpropagation step, the expected rewards from the simulation are propagated back through the selected nodes, updating their estimated values. This structured simulation and proposal-guided exploration together significantly enhance planning safety, efficiency, and realism in complex urban driving scenarios.

\subsection{Implementation Details}
We use the following configuration and hyperparameters in the following experiments.
Historical data span 1 second at 10 Hz, capturing the ego vehicle and surrounding agents.  
The planning horizon covers 3 seconds into the future, also at 10 Hz.
The occupancy prediction model operates on a $400 \times 400$ raster grid with 0.25\,m per pixel resolution.
For the multimodal ego proposal network~\cite{zhou_hivt_2022}, we follow the original configuration with an embedding dimension of 128 and a 50\,m local region radius. At each planning step, the model generates $K=3$ ego trajectory proposals, each passed to the MCTS planner for heuristic-guided exploration.

The MCTS planner performs 200 iterations per proposal to select the best trajectory. The number of child nodes per tree node is limited by the progressive widening rule in \eqref{eq:progressive_widening}, with \(k = 2.0\) and \(\gamma = 0.5\).  
Candidate controls apply perturbations of $\pm0.5\,$m/s$^2$ in acceleration and $\pm0.1\,$rad in steering around each proposal. Rollouts add zero-mean Gaussian noise with standard deviations $0.2\,$m/s$^2$ for acceleration and $0.03\,$rad for steering, respectively.

\section{Experiments}

\subsection{Datasets}
We evaluate our work on two large-scale datasets.

\noindent \textbf{nuPlan Dataset.}
Our primary evaluation is conducted on the nuPlan dataset \cite{nuplan} using two benchmark subsets. Unless otherwise noted, all nuPlan experiments follow the setup of the baseline DTPP \cite{huang2024dtpp}, as shown in Table~\ref{tab:closed_loop_nuplan}. This setting includes 10 dynamic scenario categories (excluding static ones) defined by the nuPlan planning challenge. We used 100,000 scenarios for training and 200 scenarios (20 per type) for testing, each lasting 15 seconds. Both closed-loop non-reactive (CL-NR) and reactive (CL-R) settings are evaluated.
In addition, we report results on the Val14 benchmark \cite{Dauner2023CORL}, which contains up to 100 test scenarios for each of the 14 types defined by the challenge. The results of other baselines in Table~\ref{tab:closed_loop_val14} are taken from \cite{cheng2024pluto}.

\noindent \textbf{DeepUrban Dataset:}  
In addition to nuPlan, we utilize the DeepUrban dataset \cite{selzer2024deepurban}, which is based on drone-recorded data over urban streets. This dataset captures a wide variety of challenging scenarios in densely populated urban areas, such as those in Munich. We follow the training/validation splits defined in the DeepUrban pipeline.

\subsection{Metrics}
For nuPlan, we adopt the official nuPlan evaluation metrics \cite{nuplan_metrics}, with the NR score and the R score denoting non‑reactive and reactive closed‑loop scores. For DeepUrban, we conduct open-loop planning evaluations, assessing the 3-second planned trajectories using the following metrics:

\noindent\textbf{Collision Rate:} Calculated as the percentage of trajectories where the ego vehicle's bounding box intersects with any other agent's bounding box at any time step. 

\noindent\textbf{Off-Road Percentage:} Defined as the proportion of trajectory points that lie outside the drivable lane polygons. 

\noindent\textbf{Progress Ratio:} Computed as the ratio between the cumulative distance traveled along the closest lane centerlines by planned and ground-truth trajectory. 
The ratio is capped at 1.0 to reflect the relative progress efficiency.

\subsection{Baseline Models}

We evaluate our approach against a broad set of SOTA methods that cover a wide spectrum of planning paradigms. They can be categorized into three groups:

\noindent\textbf{Rule-Based Methods:} IDM~\cite{Treiber_2000}, a classical car-following model; PDM-Closed~\cite{Dauner2023CORL}, the nuPlan competition winner combining IDM with rule-based scoring; and a manually tuned tree search planner.

\noindent\textbf{Learning-Based Methods:} CNN-based RasterModel~\cite{nuplan}, UrbanDriver~\cite{scheel2021urban} with vectorized Transformer inputs, Transformer-based PlanTF~\cite{cheng2023plantf}, MLP-based planner PDM-Open~\cite{Dauner2023CORL}, and HiVT~\cite{zhou_hivt_2022} as our proposal network.

\noindent\textbf{Hybrid Methods:} GameFormer~\cite{Huang_2023_ICCV} integrates game‑theoretic prediction; PlanTF‑H~\cite{cheng2023plantf} adds refinement to PlanTF; PLUTO~\cite{cheng2024pluto} uses contrastive learning; DiffStack~\cite{karkus2022diffstack} enables differentiable planning with modular interpretability; TPP~\cite{Chen2023TreestructuredPP} and DTPP~\cite{huang2024dtpp} employ ego‑conditioned predictions with structured search.

\subsection{Results and Analysis}

\noindent\textbf{Closed- and Open-Loop Planning.}
Table~\ref{tab:closed_loop_nuplan} shows that our method (HYPE) achieves the lowest collision rates among all methods.
It ranks second in overall scores, nearly on par with PDM-Closed~\cite{Dauner2023CORL} which is highly optimized for the nuPlan challenge metrics~\cite{huang2024dtpp}.
Compared to the best-performing ego-conditioned baseline DTPP~\cite{huang2024dtpp}, which uses centerline-guided proposal sampling and object-based prediction, HYPE reduces the reactive collision rate from 0.025 to 0.005 and improves the NR and R scores by +0.004 and +0.015, respectively.

\begin{table}[tb]
    \centering
    \caption{Planning performance on the nuPlan dynamic scenarios \cite{huang2024dtpp}. The \textbf{best} and \underline{second-best} values are highlighted.}
    \label{tab:closed_loop_nuplan}
    \renewcommand{\arraystretch}{1.2}
    \resizebox{\columnwidth}{!}{%
    \begin{tabular}{l|c c c c}
    \hline
    \textbf{Method} & \textbf{NR Score} ($\uparrow$) & \textbf{NR Collision} ($\downarrow$) & \textbf{R Score} ($\uparrow$) & \textbf{R Collision} ($\downarrow$) \\
    \hline
    Urban Driver \cite{scheel2021urban}         & 0.6482 & 0.180  & 0.6598 & 0.180  \\
    IDM \cite{Treiber_2000}                  & 0.6396 & 0.160  & 0.6168 & 0.150  \\
    TPP \cite{Chen2023TreestructuredPP}                 & 0.7388 & 0.100  & 0.7699 & 0.065 \\
    HiVT* \cite{zhou_hivt_2022}   & 0.8137 & 0.065 & 0.8011  & 0.060\\
    PDM-Closed \cite{Dauner2023CORL}                  & \textbf{0.9061} & 0.035 & \textbf{0.9150} & \underline{0.015} \\
    PLUTO \cite{cheng2024pluto}                 & 0.8835 & \underline{0.025} & 0.8796 & 0.030 \\
    DTPP \cite{huang2024dtpp}                 & 0.8964 & \underline{0.025} & 0.8978 & 0.025 \\
    
    \hline
    HYPE (ours)    & \underline{0.9002} & \textbf{0.020} & \underline{0.9127} & \textbf{0.005} \\
    \hline
    \multicolumn{4}{l}{\scriptsize{* HiVT enhanced with route embedding.}}
    \vspace{-1.5em}
    \end{tabular}%
    }
\end{table}

Table~\ref{tab:closed_loop_val14} presents the results on the nuPlan Val14 benchmark~\cite{Dauner2023CORL}, where the NR score is the main overall metric~\cite{cheng2024pluto}.
HYPE achieves the second-best R score, and outperforms all baselines in collision and TTC. The improvement over the second-best PLUTO is by +0.32 (collision) and +1.51 (TTC), and over the ego-conditioned planner DTPP by +2.49 and +4.88, respectively.
Still, it maintains the second highest comfort.
The effectiveness of the MCTS integration is illustrated by improvements in collision, TTC and R score relative to our proposal network HiVT*~\cite{zhou_hivt_2022}.
Despite slightly trading off progress and speed, HYPE retains most of the proposal efficiency while substantially enhancing safety.

Table~\ref{tab:open_loop_comparison_deepurban} presents the evaluation results on the DeepUrban dataset, showing that HYPE achieves the lowest collision rate and strong progress, outperforming DTPP across all metrics.  
Compared to HiVT*, HYPE reduces collisions by 0.013 while retaining high progress (0.906 vs. 0.919).  
It also improves progress over the purely rule-based tree search by +0.06.
These results demonstrate the benefit of combining multimodal proposal guidance with structured planning in diverse and challenging urban scenarios.

\begin{table*}[htbp]
    \centering
    \caption{Planning performance on the nuPlan Val14 benchmark \cite{cheng2024pluto}.} 
    \label{tab:closed_loop_val14}
    \resizebox{\textwidth}{!}{%
    \begin{tabular}{l|l|c c c c c c c c}
    \hline
    \textbf{Type} & \textbf{Planner} & \textbf{NR Score} ($\uparrow$) & \textbf{Collisions} ($\uparrow$) & \textbf{TTC} ($\uparrow$) & \textbf{Drivable} ($\uparrow$) & \textbf{Comfort} ($\uparrow$) & \textbf{Progress} ($\uparrow$) & \textbf{Speed} ($\uparrow$) & \textbf{R Score} ($\uparrow$) \\
    \hline
    \textcolor{gray} {Expert} &  \textcolor{gray} {Log-Replay} & \textcolor{gray} {93.68} & \textcolor{gray} {98.76} & \textcolor{gray} {94.40} & \textcolor{gray} {98.07} & \textcolor{gray} {99.27} & \textcolor{gray} {98.99} & \textcolor{gray} {96.47} & \textcolor{gray} {81.24} \\
    \hline
    {Rule-Based} & IDM \cite{Treiber_2000} & 79.31 & 90.92 & 83.49 & 94.04 & 94.40 & 86.16 & 97.33 & 79.31 \\
    & PDM-Closed \cite{Dauner2023CORL} & \underline{93.08} & 98.07 & 93.30 & \textbf{99.82} & 95.52 & 92.13 & \textbf{99.83} & \textbf{93.20} \\
    \hline
    {Pure Learning} & PDM-Open \cite{Dauner2023CORL} & 50.24 & 74.54 & 69.08 & 87.89 & \textbf{99.54} & 69.86 & 97.72 & 54.86 \\
    & RasterModel \cite{nuplan} & 66.92 & 86.97 & 81.46 & 85.04 & 81.46 & 80.60 & 98.03 & 64.66 \\
    & HiVT* \cite{zhou_hivt_2022} & 84.70 & 92.23 & 88.31 & 97.60 & 93.78 & 91.46 & 98.40 & 83.91 \\
    & PlanTF \cite{cheng2023plantf} & 85.30 & 94.13 & 90.73 & 96.79 & 93.67 & 89.83 & 97.78 & 77.07 \\
    & PLUTO$^\dagger$ (w/o post.) \cite{cheng2024pluto} & 89.04 & 96.18 & 93.28 & 98.53 & 96.41 & 89.56 & 98.13 & 80.01 \\
    \hline
    {Hybrid} & GameFormer \cite{Huang_2023_ICCV} & 82.95 & 94.32 & 86.77 & 94.87 & 93.39 & 89.04 & \underline{98.67} & 83.88 \\
    & DTPP \cite{huang2024dtpp} & 85.44 & 96.13 & 90.67 & 94.79 & 93.67 & 89.60 & 97.07 & 85.91 \\
    & PlanTF-H \cite{cheng2023plantf} & 89.96 & 97.06 & 93.38 & 97.79 & 91.08 & \underline{92.90} & 98.01 & 88.08 \\
    & PLUTO \cite{cheng2024pluto} & \textbf{93.21} & \underline{98.30} & \underline{94.04} & \underline{99.72} & 91.93 & \textbf{93.65} & 98.20 & 92.06 \\
    & HYPE (ours) & 91.04 & \textbf{98.62} & \textbf{95.55} & 96.91 & \underline{96.57} & 91.07 & 97.88 & \underline{92.63} \\
    \hline
    \multicolumn{4}{l}{\footnotesize{* HiVT enhanced with route embedding.}}
    \vspace{-0.5em}
    \end{tabular}%
    }
\end{table*}

\begin{table}[tb]
\vspace{-0.5em}
\caption{Planning performance on the DeepUrban dataset.}
\label{tab:open_loop_comparison_deepurban}
\renewcommand{\arraystretch}{1.2}  
\resizebox{\columnwidth}{!}{%
\begin{tabular}{l|c c c c}
\hline
\textbf{Method} & \textbf{Collision} ($\downarrow$) & \textbf{Off Road} ($\downarrow$) & \textbf{Progress} ($\uparrow$)  \\
\hline
DiffStack \cite{karkus2022diffstack}              & 0.021 & \textbf{0.004} & 0.866 \\
DTPP \cite{huang2024dtpp}                   & 0.017 & 0.012 & 0.895 \\
HiVT* \cite{zhou_hivt_2022}                   & 0.018 & \underline{0.009} & \textbf{0.919} \\
Tree Search    & \underline{0.007} & 0.014 & 0.846 \\
\hline
HYPE (ours)      & \textbf{0.005} & 0.011 & \underline{0.906} \\
\hline
\multicolumn{4}{l}{\scriptsize{* HiVT enhanced with route embedding.}}
\end{tabular}
}
\vspace{-0.5em}
\end{table}

\begin{figure*}[htbp]
    \centering
    \includegraphics[width=\textwidth]{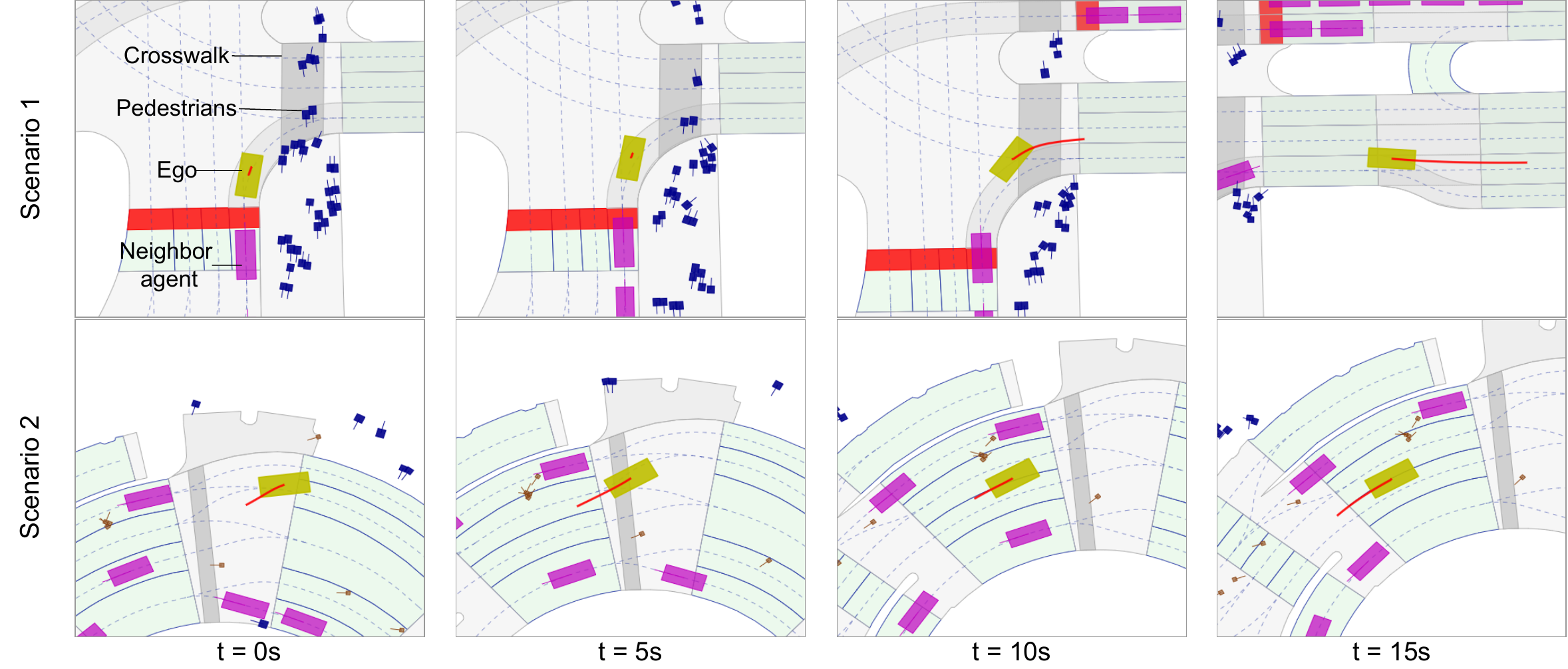}
    \vspace{-1.5em}
    \caption{Qualitative results of closed-loop planning for two representative scenarios from the nuPlan test set. Lines with color red represent the ego planned trajectory. Small brown boxes in Scenario 2 denote traffic cones. The scenarios are detailed as follows: (1) The ego vehicle halts at a stop line to yield to pedestrians crossing before a right turn, demonstrating compliance with pedestrian right-of-way.  (2) Encountering a stationary vehicle within its lane, the ego vehicle performs a lane change to bypass the obstacle, highlighting its dynamic decision-making and awareness of road topology.}
    \vspace{-0.5em}
    \label{fig:results_1}
\end{figure*}

\noindent\textbf{Qualitative Results.}
Fig.~\ref{fig:results_1} presents selected scenarios from the nuPlan test set, showcasing the robust performance of HYPE in complex, interactive urban driving situations.

\begin{figure}[tb]
    \centering
    \includegraphics[width=\linewidth]{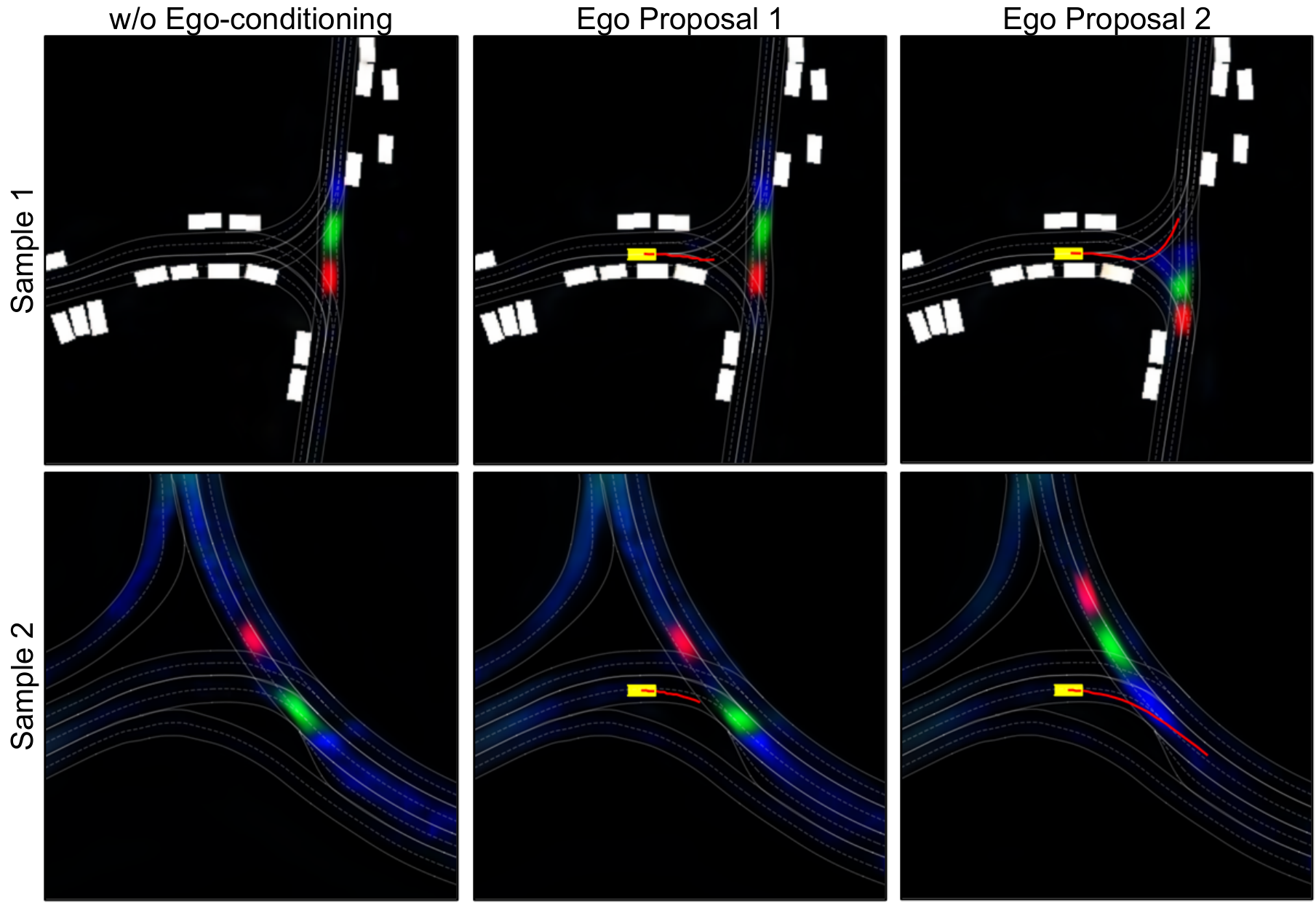}
    \vspace{-1.5em}
    \caption{Qualitative results for ego-conditioned occupancy prediction. Our model predicts occupancy grids for surrounding agents conditioned on different ego trajectory proposals. In the visualization, the ego vehicle is shown in yellow, and its trajectory proposals are in red. Occupancy predictions of other agents at future timesteps are color-coded: red for $t=1s$, green for $t=2s$, and blue for $t=3s$. White regions indicate similar occupancy probabilities across all timesteps, typically reflecting static agents.}
    \vspace{-1.0em}
    \label{fig:occ_ego_cond_comparison}
\end{figure}

\noindent\textbf{Ego-Conditioned Occupancy Predictions.}
Fig.~\ref{fig:occ_ego_cond_comparison} presents qualitative examples of ego-conditioned occupancy predictions at intersections, where the ego vehicle intends to merge onto the main road. Ego-conditioned occupancy predictions exhibit scene-consistent reasoning. Under varying ego proposals, our model predicts distinct and plausible responses from surrounding agents, enabling the cost function module to assess scenarios more accurately and help the planner generate more confident and targeted planning decisions.

\begin{figure}[tb]
    \centering
    \includegraphics[width=\linewidth]{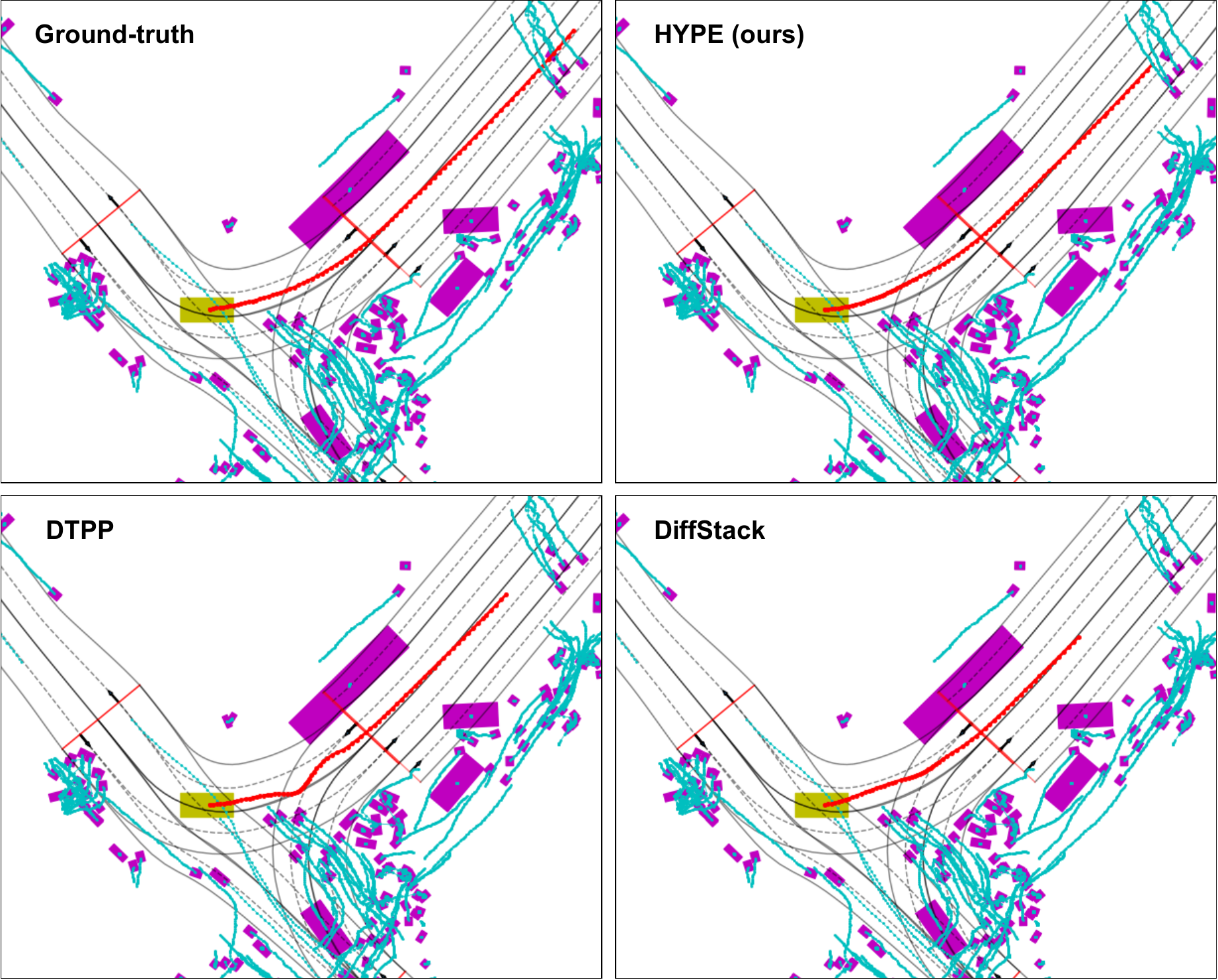}
    \vspace{-1.5em}
    \caption{Qualitative results of 8-second trajectory planning in a crowded urban scenario from DeepUrban MunichTal. Each subfigure visualizes the planned ego trajectory from different planners: HYPE (ours), DTPP, and DiffStack. The red line represents the planned future trajectory of the ego vehicle, with the yellow box indicating its current position. Magenta boxes denote surrounding agents along with their ground-truth future trajectories.} 
    \vspace{-1.0em}
    \label{fig:munich_planners_compare}
\end{figure}

\noindent\textbf{Data-Driven Ego Future Proposals.}
To demonstrate the advantage of using data-driven multimodal proposals as heuristic guidance and the effectiveness of ego-conditioned occupancy prediction, we compare the planning performance with multiple baselines in challenging DeepUrban MunichTal scenarios as shown in Fig.~\ref{fig:munich_planners_compare}.
Compared to centerline-based and goal-point proposal sampling approaches, our planner generates the smoothest trajectory and achieves the best progress in crowded and interactive scenes.  
Table~\ref{tab:ablation_deepurban_munich} further shows that combining learnable proposals that implicitly encode complex and long-term behavior reasoning with ego-conditioned occupancy prediction enables the use of a lightweight cost function (collision avoidance and proposal deviation), achieving the lowest collision and higher progress in dense urban environments.

\begin{table}[tb]
\centering
\caption{Planning performance on the DeepUrban MunichTal subset.}
\label{tab:ablation_deepurban_munich}
\renewcommand{\arraystretch}{1.2}
\begin{tabular}{l|c c c}
\hline
\textbf{Method} & \textbf{Collision} ($\downarrow$) & \textbf{Off Road} ($\downarrow$) & \textbf{Progress} ($\uparrow$) \\
\hline
DiffStack \cite{karkus2022diffstack}              & 0.033 & \textbf{0.010} & 0.738 \\
DTPP \cite{huang2024dtpp}                   & 0.024 & \underline{0.013} & {0.754} \\
HiVT* \cite{zhou_hivt_2022}                   & 0.027 & \textbf{0.010} & \textbf{0.823} \\
Tree Search         & \underline{0.015} & 0.018 & 0.731 \\
\hline
HYPE (ours)      & \textbf{0.012} & \underline{0.013} & \underline{0.781} \\
\hline
\multicolumn{4}{l}{\scriptsize{* HiVT enhanced with route embedding.}}
\end{tabular}%
\vspace{-1.8em}
\end{table}

\begin{table}[tb]
    \centering
    \small
    \caption{Ablation of route embedding and ego-conditioning on DeepUrban.}
    \label{tab:ablation_deepurban_components}
    \renewcommand{\arraystretch}{1.2}
    \resizebox{\columnwidth}{!}{%
    \begin{tabular}{c|l|ccc}
    \hline
    \textbf{Model} & \textbf{Description} & \textbf{Collision} ($\downarrow$) & \textbf{Off Road} ($\downarrow$) & \textbf{Progress} ($\uparrow$) \\
    \hline
    $M_0$ & Base & 0.007 & 0.012 & 0.853\\
    $M_1$ & $M_0$ + Route Embedding(RE) & 0.007 & 0.015 & 0.883 \\
    $M_2$ & $M_1$ + Ego-Conditioning(EC)& \textbf{0.005} & \textbf{0.011} & \textbf{0.906} \\
    \hline
    \multicolumn{4}{l}{\footnotesize Base: no conditioning, no route embedding.}
    \vspace{-1.3em}
    \end{tabular}%
    }

    \vspace{0.5cm}
    \centering
    \small
    \caption{Ablation of route embedding and ego-conditioning on nuPlan.}
    \label{tab:ablation_nuplan_components}
    \renewcommand{\arraystretch}{1.2}
    \resizebox{\columnwidth}{!}{%
    \begin{tabular}{c|l|ccc|ccc}
    \hline
    \textbf{Model} & \textbf{Description} & \multicolumn{3}{c|}{\textbf{Non-Reactive (NR)}} & \multicolumn{3}{c}{\textbf{Reactive (R)}} \\
    & & Score$\uparrow$ & Coll.$\downarrow$ & Prog.$\uparrow$ & Score$\uparrow$ & Coll.$\downarrow$ & Prog.$\uparrow$ \\
    \hline
    $M_0$ & Base & 0.8691 & \textbf{0.020} &0.8902 & 0.8736 & 0.015 &0.8821 \\
    $M_1$ & $M_0$ + RE & 0.8942 & \textbf{0.020} & 0.9091 & 0.9036 & 0.015 &0.9112 \\
    $M_2$ & $M_1$ + EC & \textbf{0.9002} & \textbf{0.020} & \textbf{0.9103} & \textbf{0.9127} & \textbf{0.005} & \textbf{0.9179} \\
    \hline
    \multicolumn{8}{l}{\footnotesize{Base: no conditioning, no route embedding.}}
    \vspace{-0.5em}
    \end{tabular}
    }
\end{table}

\begin{figure}[tb]
    \centering
    \includegraphics[width=0.8\linewidth]{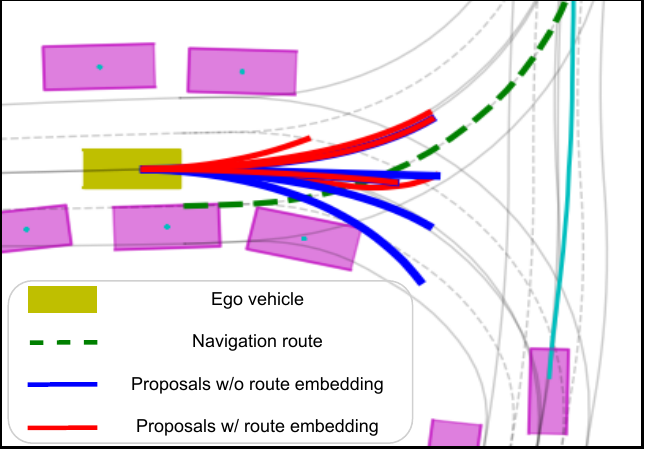}
    \vspace{-0.5em}
    \caption{Multimodal trajectory proposals generated by the proposal network, with and without route embedding.}
    \vspace{-1.0em}
    \label{fig:multi_proposals_route}
\end{figure}

\subsection{Ablation Studies}
\noindent\textbf{Ego-Conditioning and Route Embedding.} We investigate the influence of route embedding in the ego proposal network and ego-conditioning in the occupancy prediction module. The ablation study includes three model variants: a base model $M_0$ without route or ego conditioning, $M_1$ with route embedding in the branch of ego trajectory proposal, and $M_2$ with additional ego-conditioning in occupancy prediction. 

As summarized in Tables~\ref{tab:ablation_deepurban_components} and~\ref{tab:ablation_nuplan_components}, both components contribute positively to planning performance. Compared to $M_0$, the inclusion of route embedding in $M_1$ improves planning metrics across the board by better aligning proposals with the navigation path, resulting in increased progress. Building upon this, the addition of ego-conditioning in $M_2$ brings further improvements in safety and progress efficiency.

On nuPlan, as shown in Table~\ref{tab:ablation_nuplan_components}, $M_2$ achieves the highest NR and R scores while reducing the collision rate to 0.005.  
Similar trends are observed on DeepUrban in Table~\ref{tab:ablation_deepurban_components}, where ego-conditioning reduces collision risk and achieves the highest progress.  
This improvement indicates that ego-conditioning enables more efficient forward motion, reducing rear-end collisions in open-loop settings.  
Overall, results underscore the benefit of integrating route-aware proposals with ego-aware occupancy reasoning. 

We selected an intersection scenario and performed inference from the same initial position to analyze the impact of route embedding on the proposal network’s ability to capture multimodal behavior.  
As shown in Fig.~\ref{fig:multi_proposals_route}, with route information, the model generates proposals that align with the intended maneuver, e.g., making a left turn.  
Without route conditioning, proposals show greater variability and may deviate from the desired path, highlighting the importance of route-awareness for consistent and goal-aligned proposals.

\begin{table}[tb]
\centering
\caption{Ablation of the number of guidance proposals on nuPlan planning performance.}
\label{tab:nuplan_ablation_mode_nums}
\renewcommand{\arraystretch}{1.2}
\resizebox{\columnwidth}{!}{%
\begin{tabular}{l|c c c c}
\hline
\textbf{Method} & \textbf{NR Score} ($\uparrow$) & \textbf{NR Collision} ($\downarrow$) & \textbf{R Score} ($\uparrow$) & \textbf{R Collision} ($\downarrow$) \\
\hline
 \textcolor{gray}{DTPP} \cite{huang2024dtpp}                 & \textcolor{gray}{0.8964} & \textcolor{gray}{0.025} & \textcolor{gray}{0.8978} & \textcolor{gray}{0.025} \\
 \hline
One-Mode      & 0.8810           & 0.025           & 0.8880           & 0.015 \\
Two-Mode    & {0.8973}             & \textbf{0.020}       & {0.8992}          & \underline{0.010} \\
Three-Mode    & \underline{0.9002} & \textbf{0.020} & \underline{0.9127} & \textbf{0.005} \\
Six-Mode      & \textbf{0.9005} & \textbf{0.020} & \textbf{0.9133} & \textbf{0.005} \\
\hline
\end{tabular}%
}

\vspace{0.5cm}
\centering
\small
\caption{Runtime analysis under varying numbers of guidance proposals.}
\label{tab:runtime_analysis}
\renewcommand{\arraystretch}{1.2}
\resizebox{\columnwidth}{!}{%
\begin{tabular}{l|cc|cc|cc}
\hline
\textbf{Modules} & \multicolumn{2}{c|}{\textbf{One Mode (s)}} & \multicolumn{2}{c|}{\textbf{Two Modes (s)}} & \multicolumn{2}{c}{\textbf{Three Modes (s)}} \\
                                  & \textbf{w/ cond.} & \textbf{w/o cond.} & \textbf{w/ cond.} & \textbf{w/o cond.} & \textbf{w/ cond.} & \textbf{w/o cond.} \\
\hline
Ego Proposal                  & 0.073 & 0.069 & 0.077 & 0.070 & 0.079 & 0.073 \\
Occupancy Prediction       & 0.041 & 0.042 & 0.040 & 0.041 & 0.050 & 0.042 \\
Rasterization                    & 0.115 & 0.114 & 0.121 & 0.114 & 0.136 & 0.114 \\
\textbf{Prediction Total}        & \textbf{0.229} & \textbf{0.225} & \textbf{0.238} & \textbf{0.225} & \textbf{0.265} & \textbf{0.229} \\
\hline
MCTS Planning                    & 0.085 & 0.081 & 0.261 & 0.236 & 0.396 & 0.401 \\
\textbf{Total Runtime}           & \textbf{0.315} & \textbf{0.306} & \textbf{0.499} & \textbf{0.461} & \textbf{0.661} & \textbf{0.630} \\
\hline
\end{tabular}
}
\vspace{-0.7em}
\end{table}

\noindent\textbf{Number of Guidance Proposals.} We investigate the effect of the number of ego guidance proposals on planning performance and runtime efficiency. As shown in Table~\ref{tab:nuplan_ablation_mode_nums}, increasing the number of modes consistently improves overall planning scores. Notably, with only two guidance proposals, our planner already outperforms DTPP \cite{huang2024dtpp} in both NR and R Score, and significantly reducing the collision rate. Between three and six guidance modes, we observe that the collision rate remains unchanged with marginal improvement in scores.

As indicated by the runtime analysis in Table~\ref{tab:runtime_analysis}, the total computational cost increases linearly with the number of guidance modes. This is primarily due to the repeated cost calculation for each proposed mode. In particular, MCTS runtime grows as each proposal requires independent evaluation and optimization in the planning tree.

Trading off performance and computational costs, we find that using three guidance proposal modes offers the most balanced compromise. It achieves near-optimal performance while mitigating the computational cost compared to higher-mode settings.


\section{Conclusion}
In this work, we present HYPE, a hyrid planning approach that integrates learned multimodal ego proposals and ego-conditioned occupancy prediction with a heuristic-guided MCTS planner. Our method addresses two key challenges in AV motion planning: ensuring safety under uncertain multi-agent interactions by providing an expressive and consistent occupancy representation explicitly conditioned on ego trajectory proposals; simplifying and generalizing the cost function design by leveraging multimodal ego proposals as medium-to-long-term heuristic guidance, enabling effective search without extensive manual tuning. 
Our experiments demonstrate that the proposed method outperforms the baseline methods on multiple benchmarks, with notable improvements in collision avoidance. 

\noindent\textbf{Limitation and Future Work. }
Despite these advantages, our current implementation still exhibits limitations in runtime efficiency. Sequential execution of the prediction and planning modules results in an overall inference time of approximately 0.66~s. However, we note that this runtime can be significantly reduced through system-level optimizations, such as GPU-accelerated rasterization, which has been shown to run within 5~ms~\cite{PredictionNet}.
Looking ahead, we plan to explore extending our approach with more advanced trajectory proposal networks, including foundation models such as vision-language models, which may offer broader semantic understanding for high-level goal conditioning.

\addtolength{\textheight}{-7.2cm}   





\section*{ACKNOWLEDGMENT}
We thank Sajad Marvi, Chun-Peng Chang, Peizheng Li, and Shuxiao Ding for insightful discussions.

\bibliographystyle{IEEEtran}
\bibliography{IEEEabrv, HIUVT, references_zotero, related_work}

\end{document}